\documentclass[a4paper,conference]{IEEEtran}

\usepackage{cite}
\usepackage[pdftex]{graphicx}
\graphicspath{{figures/}}
\usepackage{amsmath}
\usepackage{color}
\usepackage{tabularx}

\begin{document}

\title{Developing Motion Code Embedding\\ for Action Recognition in Videos}

\author{
  \IEEEauthorblockN{Maxat Alibayev, David Paulius, and Yu Sun}
  \IEEEauthorblockA{
    Department of Computer Science and Engineering 
    \\
    University of South Florida
    \\
    Tampa, Florida, United States of America\\
    Email: yusun@usf.edu
  }
}

\maketitle

\begin{abstract}
 In this work, we propose a motion embedding strategy known as motion codes, which is a vectorized representation of motions based on a manipulation's salient mechanical attributes.
 These motion codes provide a robust motion representation, and they are obtained using a hierarchy of features called the motion taxonomy. 
 We developed and trained a deep neural network model that combines visual and semantic features to identify the features found in our motion taxonomy to embed or annotate videos with motion codes. To demonstrate the potential of motion codes as features for machine learning tasks, we integrated the extracted features from the motion embedding model into the current state-of-the-art action recognition model. The obtained model achieved higher accuracy than the baseline model for the verb classification task on egocentric videos from the EPIC-KITCHENS dataset.
\end{abstract}

\IEEEpeerreviewmaketitle

\section{Introduction}

Verb classification in videos is one of two major sub-tasks of action 
recognition. Recently researchers have acquired impressive results with deep learning models. Unlike object detection and classification, verb classification heavily relies on detecting motions, so current models often use pre-computed optical flow vectors of the videos. However, there is still a big gap between the visual features from videos, including motion features from optical flow vectors, and their semantic labels. Besides, the feature maps provided by deep ConvNets are not explainable for a human to draw any conclusions and can only be appropriately interpreted by neural networks when provided with a large enough training dataset.

Using verbs as labels for classification also has several drawbacks. 
Firstly, verbs can be too ambiguous or insufficiently detailed to describe actions in demonstrations; instead, multiple words or phrases may be required to represent motions. 
In addition to that, verbs cannot be easily measured against one another. 
Hence, distances between such verbs in a language space do not equate to the reality of manipulation.
Conventional word representation methods, such as one-hot encoded vectors or Word2Vec \cite{mikolov2013distributed}, do not consider physical motions.
Instead, they focus on representing word semantics based on the verb class that a word represents or the context it appears within the text corpora.
As a result, since there is no physical or mechanical meaning behind these vectorized representations, two semantically different verbs may display similar motions, while the same word may correspond to multiple motions that are visually different.
Therefore, the loss functions defined solely by verb labels are not sufficient for models that classify actions in videos.
These points show that a new representation of actions is required in addition to the semantic labels (viz. verbs and nouns) that can represent visually salient motion features in a concise way.

In this work, we introduce motion embedding using \textit{motion codes}, which are a vectorized representation of manipulations in an attribute space that describes mechanical and physical features, as a solution to the disparity between representations.
Such features mainly fall under trajectory and contact descriptors.
Motion codes are created using a hierarchy of features akin to a decision tree known as the \textit{motion taxonomy}, which we introduce in Section \ref{sec:motion_taxonomy}.
The motion taxonomy outlines five major components that are considered when building these codes.
A classifier or model, such as a deep neural network used in this paper, can be trained to predict these features separately, thus allowing us to embed motion features in an unsupervised fashion; in Section \ref{sec:method}, we propose a prediction model that derives a motion code for each demonstration video in this way.
We evaluate the performance of this prediction model in Section
~\ref{sec:results} and show how these features can improve verb classification accuracy.

\section{Related Works}
\label{sec:related-work}
\subsection{Action Recognition}

Currently there are numerous deep learning models designed for the purpose of action recognition that use convolutional and recurrent neural networks~\cite{karpathy2014large,simonyan2014two,wang2015towards,wang2016tsn,feichtenhofer2016convolutional,bilen2016dynamic,donahue2015long,ullah2017action,varol2017long,tran2015learning,carreira2017quo,wang2018video}. These methods have achieved state-of-the-art results on well-known datasets such as UCF101 \cite{soomro2012ucf101} and HMDB51 \cite{kuehne2011hmdb}.  

The work in \cite{karpathy2014large} presented the idea to use ConvNets for video classification. They studied different ways of combining spatial information from video frames, introducing early-fusion, late-fusion, and slow-fusion strategies. From this, \cite{simonyan2014two,wang2015towards,feichtenhofer2016convolutional,wang2016tsn} used a two-stream approach, where two separate ConvNets were trained on two input modalities, namely an RGB frame and stacked optical flow vectors. In all cases, the outputs of two models were late fused, with the exception of the model from \cite{feichtenhofer2016convolutional} that fused the modalities at the early convolutional layers. An alternative solution in \cite{bilen2016dynamic} summarized the entire video sequences into a single image and used conventional image classification models.

One shortcoming of the methods above is that they do not adequately leverage temporal information of the videos from RGB frames and mostly delegate this task to optical flow frames. This issue also limits their capability to classify longer demonstrations of action; only the TSN model, proposed in \cite{wang2016tsn}, was able to handle that via a sparse temporal sampling method. Another solution to this problem was to use recurrent neural networks to preserve the temporal ordering of the frames, which was done in \cite{donahue2015long,varol2017long,ullah2017action} with LSTM and 
bidirectional LSTM layers. 
The experiments from \cite{varol2017long} also highlight the importance of accurate motion estimation, which is one of the objectives of our work. 

Other video action recognition methods are based on adding a temporal dimension to the ConvNets. The first successful implementation of this method was done in \cite{tran2015learning}, where they showed that 3D convolutional layers are more suitable for preserving temporal information. The later work in \cite{carreira2017quo} introduced two-stream inflated 3D ConvNets and demonstrates significant improvements in action recognition on both the UCF101 and HMDB datasets. The main contribution of this work was the idea of inflating 2D image classification networks with temporal dimension and pre-training it on a new large-scale Kinetics dataset, which contains 400 activity categories with over 400 videos in each category.

All methods mentioned above established the state-of-the-art performances on action recognition tasks. 
It is important to note that all methods highly benefit from a good estimation of optical flow vectors, which are supposed to represent information about the motion in the video. However, in addition to RGB frames, optical flow vectors as features are too low-level for this problem. Therefore, our work focuses on predicting a higher-level motion representation using our motion taxonomy (shown as Figure~\ref{fig:taxonomy}). 
As elaborated in~\cite{varol2017long}, motion feature estimation is important for action recognition, and the results of the application of our proposed motion code embedding model on verb classifier support this idea.

\subsection{Motion Embedding}

The idea of applying embedding on a visual content was used in image classification via attribute-based label embedding \cite{akata2015label}, as well as in multi-modal retrieval via embedding visual and semantic data into a common vector space \cite{wray2019fine_pos,hahn2019action2vec}. Our work shares some similarities with these methods. We embed motions in the videos by assigning proper values to the set of motion components from motion taxonomy, which is similar to the idea of using a finite set of attributes from \cite{akata2015label}. We also extracted higher-level visual features via ConvNets and used them for further embedding as in \cite{wray2019fine_pos,hahn2019action2vec}. However, instead of embedding features into some arbitrary space, we extract low-dimensional motion codes representing motions in the video, thus obtaining more explainable embedding vectors. In other words, vectors can be used to explain what happens in the motion in a way that is relevant for robotic manipulation as well as contrasting between two motions.

In works such as 
\cite{daruna2019robocse,DBLP:journals/corr/FuldaRMW17,DBLP:journals/corr/abs-1809-00589,Fang_2018_CVPR,Chao_2015_CVPR}, the idea of embedding wass used to derive or draw relationships in a vectorized space for affordance \cite{Gibson_1977}. Semantic vectors can be more descriptive and suitable to answer questions such as object similarity and outcome of actions. In the case of \cite{Fang_2018_CVPR}, the researchers aimed to draw these details of the region of interest and label directly from demonstration videos. Compared to our work, motion codes will more vividly describe actions to extend beyond language labels that were typically used to represent or describe affordance \cite{paulius2019survey, paulius2016functional, paulius2018functional}. 

\section{Motion Taxonomy}
\label{sec:motion_taxonomy}

\begin{figure}
  \centering
  \includegraphics[trim={0.7cm 0.7cm 0.7cm 0.7cm},clip,width=0.9\columnwidth]{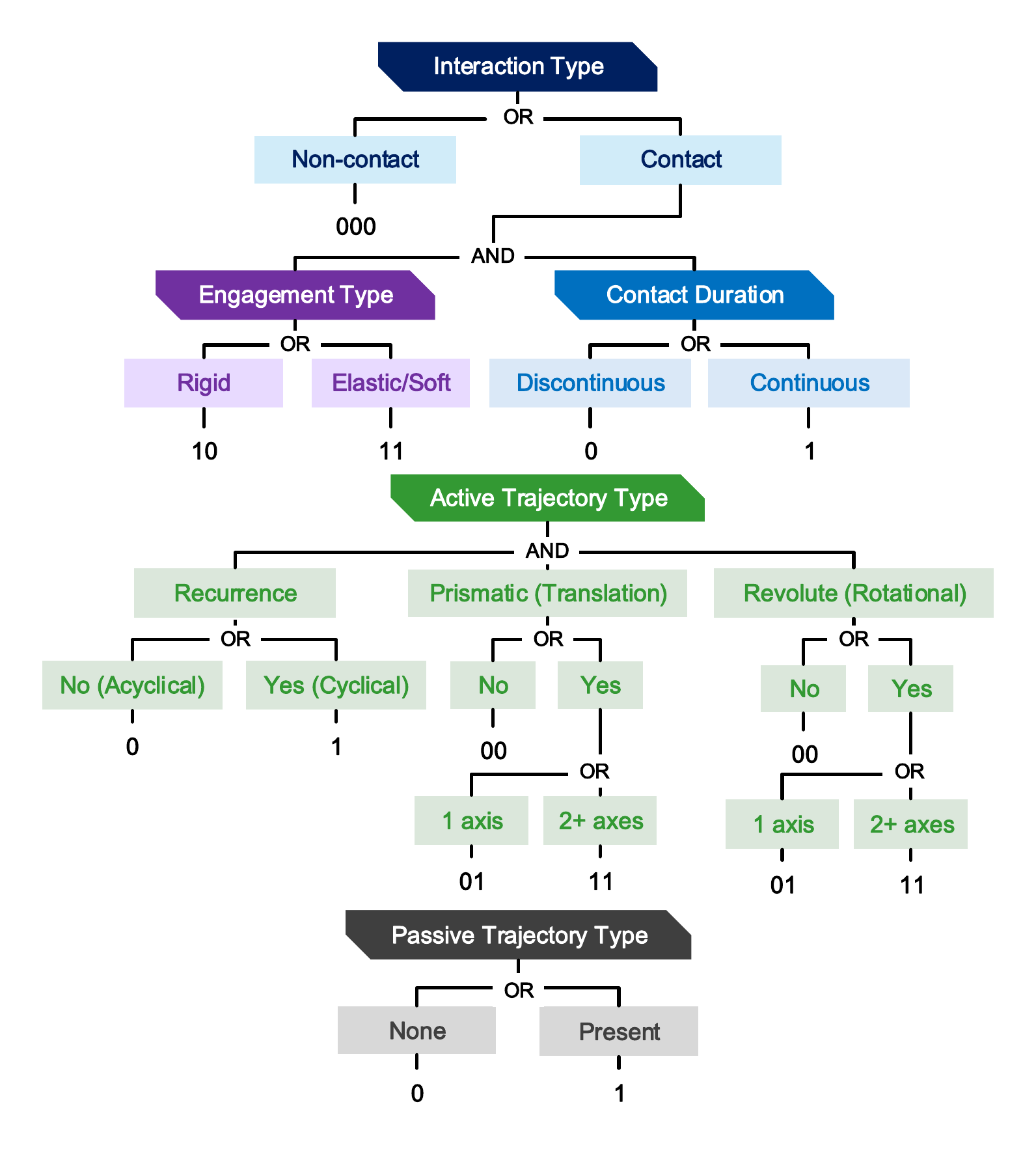}
  \caption{
    Illustration of our motion taxonomy. 
    A motion code is formed by appending bits for contact features, the active object's trajectory, and the passive object's motion with respect to active object by following the tree.
  }
  \label{fig:taxonomy}
\end{figure}

We define motion code embedding using the motion taxonomy \cite{paulius2019manipulation,paulius2020taxonomy, alibayev2020Estimating}. 
The motion taxonomy is a hierarchical categorization of physical and mechanical attributes that can be used to vividly describe manipulations in a manner that is useful to a robot.
It incorporates several features, such as contact type and motion trajectory, that can be used to visually distinguish motions from one another, which cannot be done using natural language. Overall, we have five components that form a motion code, which are each represented as substrings: \textit{interaction type}, \textit{active trajectory recurrence}, \textit{active prismatic trajectory}, \textit{active revolute trajectory}, and \textit{passive trajectory type}.
All substrings are combined together to form a single motion code.
We illustrate a flowchart of the taxonomy featuring these components as Figure \ref{fig:taxonomy}.

When creating motion codes, it is important to note \textit{active} and \textit{passive} objects of action since many actions involve a pair of objects \cite{sun2014object, ren2013human, sun2015modeling}. We define an active object as either the hand or the combination of a hand and tool/utensil that acts upon a passive object. A passive object is defined as any object that is being manipulated or acted upon by the active object. Motion codes are designed to describe a manipulation based on these concepts. 

\subsection{Describing Interaction}

Any manipulation can be identified as either \textit{contact} or 
\textit{non-contact} interaction.
A contact manipulation means that contact is established between active and passive objects-in-action; on the contrary, a non-contact manipulation means that during the action, there is little to no contact made between the active and passive objects-in-action.
Should a manipulation be identified as a contact manipulation, it can be further described by its \textit{engagement type} and the \textit{duration of contact} made.

The engagement type of a motion describes whether the action causes any change or deformation of state, shape, or structure among the objects-in-action or not.
As such, we have two cases: \textit{rigid} engagement and \textit{soft} engagement.
In a rigid engagement motion, we do not observe any deformation; an example of such a motion is picking or placing an object since neither the active hand object nor the passive object being moved will change in its structure.
In a soft or elastic engagement motion, we observe deformation in either the active or passive object's structure; an example of this is peeling, where the passive object being peeled by an active peeler object will physically change.

Contact duration describes how long contact persistently remains between the active and passive objects.
We can either observe \textit{continuous} or \textit{discontinuous} contact.
When assigning this bit, one can determine a threshold for the amount of time that is considered as continuous contact.
In our work, we consider contact to be continuous when contact persists for about $\sim$80\% of the manipulation action.

\subsection{Describing Trajectory}

In robotics, motion trajectory is an important component to manipulation \cite{huang2019dataset, huang2016recent, huang2015generating}.
For this reason, it is important to define manipulations while considering the trajectory of the end-effector and objects-in-action.
In describing the active object-in-action trajectory, we mainly identify whether the trajectory exhibits prismatic or revolute trajectory characteristics.

A motion's trajectory is \textit{prismatic} if the active object is moved about or translated about an axis or place; specifically, we can have 1-dimensional (moving across a single axis), 2-dimensional (moving across a plane), or 3-dimensional (moving across a manifold space) prismatic trajectory.
This is akin to having 1 to 3 degrees of freedom (DOF).
The term prismatic refers to the prismatic joints used in robotics.
A prismatic trajectory can be observed in motions such as cutting, peeling, or pushing.
On the other hand, a motion's trajectory is \textit{revolute} if the active object's orientation is changed as needed for the manipulation; in other words, it is rotational motion.
The term ''revolute'' refers to the revolute joints that a robot would rely upon to perform rotational motions.
As with prismatic trajectory, a revolute trajectory can range from 1 to 3 DOF.
A revolute trajectory can be observed in motions such as pouring or flipping.
Together, prismatic and revolute trajectories describe position and orientation change of motion.
In an action, we may observe only prismatic motion, only revolute motion, both prismatic and revolute motion, or none, as these two types are not mutually exclusive.
For our experiments, we simplified the definition of trajectory to have zero (\textit{`00'}), one (\textit{`01'}), or many DOF (\textit{`11'}), since we are using egocentric videos.

When describing a trajectory with the passive object(s) in action, we identify whether or not the passive object moves with respect to the active object.
We opted to do this as passive objects may simply move along with the active object (such as the case with a hand gripping an object); furthermore, it is already a challenge to automatically extract detailed trajectory features from 
demonstration videos.

Finally, in a motion code, we can describe if there is a recurrence or any repetition in trajectory.
We describe a motion as acyclical or cyclical.
This may be important to note if a motion is not complete, or more importantly, if the robot did not successfully execute the motion.
Motions such as mixing or stirring will typically exhibit repetition.

\begin{table}
  \setlength{\tabcolsep}{4pt}
  \footnotesize
  \centering
  \caption{
    Motion codes for common manipulations based on Figure \ref{fig:taxonomy}.
    Bits are separated by hyphens with respect to its component.
  }
  \label{tab:motion_code}

  \begin{tabularx}{0.48\textwidth}{lX}
    \hline\noalign{\smallskip}
    Motion Code & Motion Verbs \\ 
    \noalign{\smallskip}
    \hline\noalign{\smallskip}

    \textit{{000}-{0-00-01}-{1}}	&	pour \\
    \textit{{000}-{1-01-00}-{1}}	&	sprinkle \\
    \textit{{100}-{0-01-00}-{0}}	&	poke, press (button), tap, adjust (button) \\
    \textit{{101}-{0-00-00}-{0}}	&	grasp, hold \\
    \textit{{101}-{0-00-01}-{0}}	&	open/close (jar), rotate, turn (knob), twist \\
    \textit{{101}-{0-01-00}-{0}}	&	spread, wipe, move, push (rigid) \\
    \textit{{101}-{0-01-01}-{0}}	&	flip \\
    \textit{{101}-{0-11-00}-{1}}	&	open/close (door) \\
    \textit{{101}-{1-00-01}-{0}}	&	shake (revolute) \\
    \textit{{101}-{1-01-00}-{0}}	&	shake (prismatic) \\
    \textit{{110}-{0-01-01}-{0}}	&	scoop \\
    \textit{{110}-{0-01-00}-{0}}	&	crack (egg) \\
    \textit{{111}-{0-01-00}-{0}}	&	insert, pierce \\
    \textit{{111}-{0-00-00}-{0}}	&	squeeze (in hand, elastic) \\
    \textit{{111}-{0-01-01}-{0}}	&	fold, unwrap, wrap \\
    \textit{{111}-{1-11-00}-{1}}	&	beat, mix, stir (liquid) \\
    \textit{{111}-{0-00-00}-{0}}	&	squeeze (in hand, rigid) \\
    \textit{{111}-{0-01-00}-{1}}	&	flatten, press, squeeze, pull apart, peel, 
                                    chop, cut, mash, peel, scrape, shave, slice \\
    \textit{{111}-{0-01-00}-{0}}	&	roll \\
    \textit{{111}-{0-11-00}-{1}}	&	saw, cut (2D), slice (2D) \\
    \textit{{111}-{1-11-00}-{1}}	&	beat, mix, stir \\
    \textit{{111}-{0-01-00}-{1}}	&	brush, sweep, spread (brush) \\
    \textit{{111}-{0-11-00}-{1}}	&	brush, sweep (surface) \\
    \textit{{111}-{0-00-00}-{1}}	&	grate \\
    \noalign{\smallskip}\hline
  \end{tabularx}
\end{table}

\begin{figure*}
  \centering
  \includegraphics[width=0.7\textwidth]{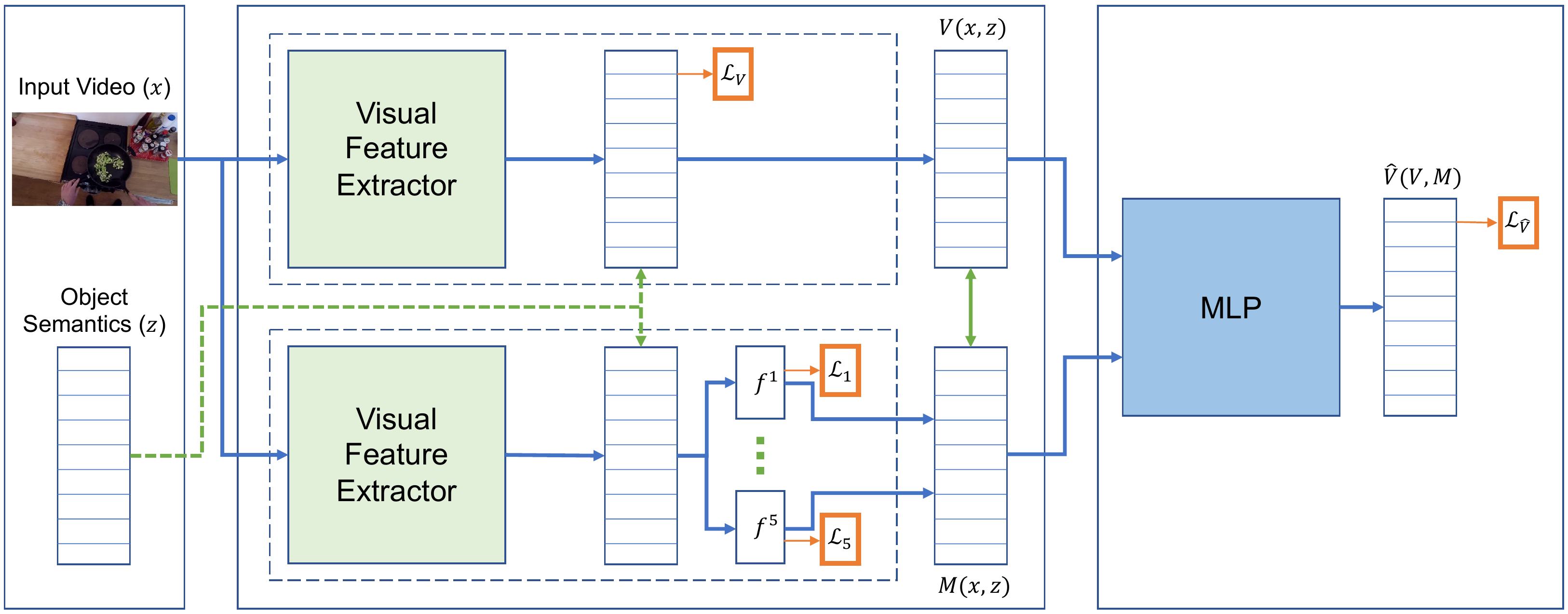}
  \caption{The structure of our verb classification framework augmented by motion code predictor.}
  \label{fig:architecture}
\end{figure*}

\subsection{Building a Motion Code}
With the taxonomy structure in Figure \ref{fig:taxonomy} as a reference, one can assign motion codes to demonstrations in household tasks, such as cooking or assembly.
Let us assume a scenario of a person chopping a cucumber with a knife; here, 
the active object-in-action is the hand/knife combination, while the passive 
object-in-action is the cucumber.
The idea is to identify the characteristics of the five components found in the 
flowchart.
Since the knife must engage in contact with the cucumber to chop it into smaller 
units, this motion would be classified as a contact manipulation of soft 
engagement type (\textit{`11'}).
This contact may be considered as continuous contact since the knife must 
contact and touch the cucumber (\textit{`1'}).

Next, we go on to describe the trajectory of the motion.
First, with the active object, the knife is usually translated in a single 
direction in order to make slices or to divide the cucumber into portions.
In cutting, there is typically no observable rotation, as slices are made evenly throughout the action.
Therefore, we will have the bits \textit{`01-00'} for prismatic and revolute 
trajectory respectively.
If we observe a demonstration where the knife is repeatedly chopping, then we 
will say that there is recurrence (\textit{`1-01-00'}); if not, then we will say 
that there is no recurrence (\textit{`0-01-00'}).
In terms of passive object motion, since the knife is moving while the cucumber is 
stationary, the cucumber is moving with respect to the active object, and so we 
will indicate this motion with \textit{`1'}.
In all, we will have a 9-bit motion code such as \textit{`111-0-01-00-1'} 
describing a chopping demonstration.

Motion codes for other common cooking or household actions are featured in 
Table \ref{tab:motion_code}.
It is important to note that there is a many-to-many mapping between the motion
codes and verbs. 
A single motion code can describe more than one verb, while one verb can 
be executed as different motions. 
For instance, in the example of cutting a cucumber, we described one scenario
where the actor is moving the knife perfectly along a single line. 
Hence, the motion has a prismatic trajectory with 1 DOF (\textit{`0-01-00'}).
However, Table \ref{tab:motion_code} shows an example where the verb \textit{cut}
has prismatic trajectory with many DOF (\textit{`0-11-00'}).
Indeed, we cannot say whether there are 1 or 2 DOF in this action since it only
depends on how the action was executed in the video.
This example demonstrates how motion codes complement the semantic information that struggles to resolve the action ambiguity by offering a more visual perspective.
Additionally, the motion code distances between manipulations from Table \ref{tab:motion_code} reflect differences in mechanics and thus benefit from explainability.

\section{Verb Classification Using Motion Code Embedding}
\label{sec:method}

In our work, we propose a model for verb classification that is described as follows.
Let $x_{i} \in \mathcal{X}$ be the $i^{th}$ video in the set of videos and $y_{i} \in \mathcal{Y}$ be the corresponding verb class label of that video. 
Conventional classification models first extract the visual features from the 
video into a latent space as $V : \mathcal{X} \to \Omega$. 
The returned feature vector can then be used as the probability distribution of 
verb classes, where the class with the highest probability value is picked. 
Our method augments that model with a motion code embedding model, which takes 
the same input video $x_{i}$ and similarly extracts its features into another latent 
space as $\theta : \mathcal{X} \to \Gamma$. 
The feature vector is then passed through five classifiers, $f^{k}: \Gamma \to \Lambda^{k}$ 
(where $k$ ranges from 1 to 5), which are responsible for classification of 
individual motion components, as we have described in Section \ref{sec:motion_taxonomy}. 
We concatenate the output probability distributions of all motion components
into a single vector, which represents a continuous motion embedding for the 
input video.
Finally, the probability distribution vector for verb classes and motion embedding 
vector are combined into a single feature vector. 
That vector is then passed through a multi-layer perceptron (MLP), denoted as 
$\hat{V}: (\Omega, \Lambda) \to \Omega$, that outputs the final verb class 
probability distribution, from which the verb class $\hat{y}_{i}$ is inferred. 
The overall structure of the network is shown in Figure \ref{fig:architecture}.
All objective functions are cross-entropy losses. 
We pre-trained the motion code embedding model separately before integrating it into 
action recognition model.
The objective function of the motion code embedding model is defined as a 
linear combination of individual losses:

\begin{align*}
  \mathcal{L}_{M} &= -\sum_{k=1}^{5}\sum_{l=1}^{C_{k}} \lambda_{k} m^{k}_{l} \log(f^{k}_{l}(x))
\end{align*}
where $\lambda_{k}$ is a constant weight and $m^{k}_{l}$ is the $l^{th}$ element 
of the ground truth one-hot vector for the $k^{th}$ motion code component. 
In our experiments, we set the value of all $\lambda_{k}$ to 1.
However, these values can be tuned to emphasize motion components that are considered more important than others.
For simplicity, we refer to the entire motion code embedding model as $M : \mathcal{X} \to \Lambda$.

We also wanted to see the impact of incorporating the semantic information about the objects-in-action. 
Verbs by themselves can be very ambiguous from a mechanical perspective because 
the same verb can be executed via different types of motions. 
For instance, one may associate the verb \textit{open} with one motion when 
opening a door and with another motion when opening a bag. 
The verbs in these cases are almost semantically identical, but visually and 
mechanically they are quite different. 
This ambiguity can be reduced with the knowledge about the object that the person is interacting with. 
Hence, such extra information can potentially benefit the motion code embedding 
model, which may benefit the verb classification accuracy. 

Let $z_{i} \in \mathcal{Z}$ be a semantic feature of the object that is being 
manipulated in the video. 
We modify our model mentioned above by combining that information with the 
visual features of the motion embedding model as shown in Figure \ref{fig:architecture}. 
For completeness, we also integrated object semantics to the baseline verb classifier.
Due to this modification, we denote the models that integrate object semantic features 
as $V(x,z)$ and $M(x,z)$.
Hence, our final model has 4 variations, namely $\hat{V}(V_{x}, M_{x})$, 
$\hat{V}(V_{x,z}, M_{x})$, $\hat{V}(V_{x}, M_{x,z})$, and $\hat{V}(V_{x,z}, M_{x,z})$.
Note that if only one model utilizes that information, then the performance of the other model will not be affected.
For instance, in model $\hat{V}(V_{x}, M_{x,z})$, the model $V_{x}$ is agnostic to
the object information, while the performance of the motion embedding model gets the direct impact.
Consequently, the performance of $\hat{V}(V_{x}, M_{x,z})$ is not directly affected by
the object knowledge neither; instead, its performance changes based on the new 
performance of motion embedding.

\section{Experiments}
\label{sec:experiments}

\subsection{Dataset Annotation}

We use the EPIC-KITCHENS \cite{Damen2018EPICKITCHENS} dataset to test how 
motion code embedding may benefit verb classification models. 
This dataset contains egocentric cooking videos that are labeled with verbs and nouns. 
We annotated 3,528 videos out of 28,472 in the training set of EPIC-KITCHENS. 
Each video was annotated with a ground truth motion code by following the 
motion taxonomy in Figure \ref{fig:taxonomy}. 
We were unable to annotate the entire dataset, as the manual labeling of videos with motion codes proved to be a very time-intensive task.
The resulting dataset contains videos with 32 unique motion codes and 33 verb classes, which was split into 2,742 videos for the training set and 786 videos for the validation set. 
We sampled 1,517 testing videos labeled with 33 verb classes from the remainder of the EPIC-KITCHENS.

\subsection{Implementation Details}
We used Two-Stream Inflated 3D ConvNet (I3D) from \cite{carreira2017quo} to extract 
visual features for both motion code embedding and baseline verb classification 
models. 
The model was pre-trained on Kinetics dataset \cite{carreira2017quo} and uses 
InceptionV1 \cite{szegedy2015going} that was pre-trained on the ImageNet 
dataset \cite{imagenet_cvpr09} as the base network. 
Both RGB and optical flow frames were used to tune the model. 
The outputs of two modalities were averaged with late-fusion. 
We chose this model because it achieved the best results on the verb 
classification task in the latest EPIC-KITCHENS Action Recognition Challenge 
2019 \cite{wang2019baidu}. 
The extracted latent feature vector is then concatenated with 300-dimensional 
object semantic vectors. These semantic vectors were obtained from Word2Vec model that was pre-trained on Google News dataset \cite{mikolov2013distributed} 
with 3 million words. 
Videos in EPIC-KITCHENS are annotated with nouns that represent the object-in-action, which are fed into Word2Vec model.

The input videos were sampled to 6 frames per second. 
During training, 12 consecutive frames were sampled from a random starting frame and sampled frames were randomly cropped and horizontally flipped. During validation and testing, all frames were used and center cropped.  This strategy was adapted from NTU-CML-MiRA team's implementation for the EPIC-KITCHENS Action Recognition Challenge 2019.

Both motion code embedding and baseline verb classifier were trained for 50 epochs with an initial learning rate of 0.0003 that reduced by 40\% every 5 epochs. The weights of the convolutional layers were frozen for the first 3 epochs to allow the top layers to fine-tune and to be properly initialized.

For the joint verb classification, $\hat{V}(V,M)$, we concatenated the outputs of both the motion code embedding and baseline verb classifier into a single input vector. The vector is then passed through an MLP with 2 fully connected layers. They were trained separately for 200 epochs with a learning rate of 0.0005. All models were trained with Adam optimizer. 

\section{Results}
\label{sec:results}

\subsection{Quantitative Results}

Before training the verb classifier, we train and evaluate the motion code 
embedding model.
Table \ref{tab:motion_prediction} shows the top-1 accuracy of predicting the 
motion codes from the video ($M_{x}$), as well as with the knowledge about the
nouns ($M_{x,z}$).
We can see that with the object semantic data, the model achieves much higher accuracy. 
It seems that the model learns the correlation between certain objects and the motions that are usually executed on them.
Given the fact that there are 180 valid motion codes and less than 3,000 training 
videos, the accuracy of almost 40\% is an acceptable result.

\setlength{\tabcolsep}{4pt}
\begin{table}
  \centering
  \caption{Motion code embedding accuracy results on validation videos (as \%).}
  \label{tab:motion_prediction}
  \begin{tabular}{llll}
  \hline\noalign{\smallskip}
    Models & RGB & Flow & Fused\\ 
    \noalign{\smallskip}
    \hline\hline
    \noalign{\smallskip}
    Baseline, $M_{x}$ & 35.1 & 35.2 & 38.9 \\
    Nouns, $M_{x,z}$ & 45.3 & 46.1 & 48.0 \\
    \noalign{\smallskip}
    \hline  
  \end{tabular}
\end{table}

With the pre-trained motion embedding model, we compared our proposed model with 
two variations of baseline I3D model, namely $V(x)$ and $V(x, z)$. 
Each baseline is compared to the models that augment it with two variations of 
the motion code embedding model, namely $M(x)$ and $M(x, z)$. 
We used top-1 verb classification accuracy as the evaluation metric for our experiments. 

\setlength{\tabcolsep}{4pt}
\begin{table}
  \centering
  \caption{
    Verb classification comparison on testing videos based on top-1 accuracy (as \%).
  }
  \label{table:verb_clasification_small}
  \begin{tabular}{llll}
  \hline\noalign{\smallskip}
    Methods & RGB & Flow & Fused \\
    \noalign{\smallskip}
    \hline\hline
    \noalign{\smallskip}
    Baseline, $V_{x}$ & 33.36 & 31.64 & 36.12 \\\\
    Motions, $\hat{V}(V_{x},M_{x})$ & 33.62 & 32.30 & 36.78\\
    Motions with nouns, $\hat{V}(V_{x},M_{x,z})$ & \textbf{34.08} & \textbf{34.74} & \textbf{38.04}\\
    \noalign{\smallskip}
    \hline
    \noalign{\smallskip}
    Baseline with nouns, $V_{x,z}$ & 38.69 & 38.76 & 41.73 \\\\
    Motions, $\hat{V}(V_{x,z},M_{x})$ & \textbf{38.89} & 37.05 & 42.06 \\
    Motions with nouns, $\hat{V}(V_{x,z},M_{x,z})$ & 38.83 & \textbf{39.95} & \textbf{42.12} \\
    \noalign{\smallskip}
    \hline
  \end{tabular}
\end{table}
\setlength{\tabcolsep}{1.4pt}

\setlength{\tabcolsep}{4pt}
\begin{table}[t]
  \centering
  \caption{
    Verb classification comparison on validation videos based on top-1 accuracy (as \%).
  }
  \label{table:verb_clasification_validation}
  \begin{tabular}{llll}
    \hline\noalign{\smallskip}
    Methods & RGB & Flow & Fused \\
    \noalign{\smallskip}
    \hline\hline
    \noalign{\smallskip}
    Baseline, $V_{x}$ & 41.60 & 39.82 & 45.04 \\
    Baseline with nouns, $V_{x,z}$ & 48.22 & 44.15 & 49.24 \\\\
    Predicted Motions, $\hat{V}(V_{x},M_{x})$ & 41.22 & 40.46 & 46.18 \\
    Predicted Motions with nouns, $\hat{V}(V_{x},M_{x,z})$ & 43.13 & 42.11 & 47.20 \\
    Ground Truth Motions*, $\hat{V}(V_{x},\Bar{M}_{x})$ & \textbf{53.82} & \textbf{53.69} & \textbf{57.63}\\
    (Using true motion code as embedding)\\
    \noalign{\smallskip}
    \hline
  \end{tabular}
\end{table}
\setlength{\tabcolsep}{1.4pt}

Table \ref{table:verb_clasification_small} outlines the results on the test set.
As one can observe from the first 2 rows, the application of motion embedding features
slightly improves the accuracy over the baseline.
However, when we augmented the motion embedding model with nouns, which previously
resulted in more accurate motion code prediction, the impact of our model becomes
more significant.
We can observe a similar trend with the second baseline model $V_{x,z}$. 
The model showed a significant boost over $V_{x}$ due to the nouns.
Nevertheless, if we apply the vanilla motion embedding model features $M_{x}$, we still manage to get higher accuracy, while more accurate motion features from $M_{x,z}$ result in even more boost in performance.

\begin{figure}
  \centering
  \includegraphics[width=0.8\columnwidth]{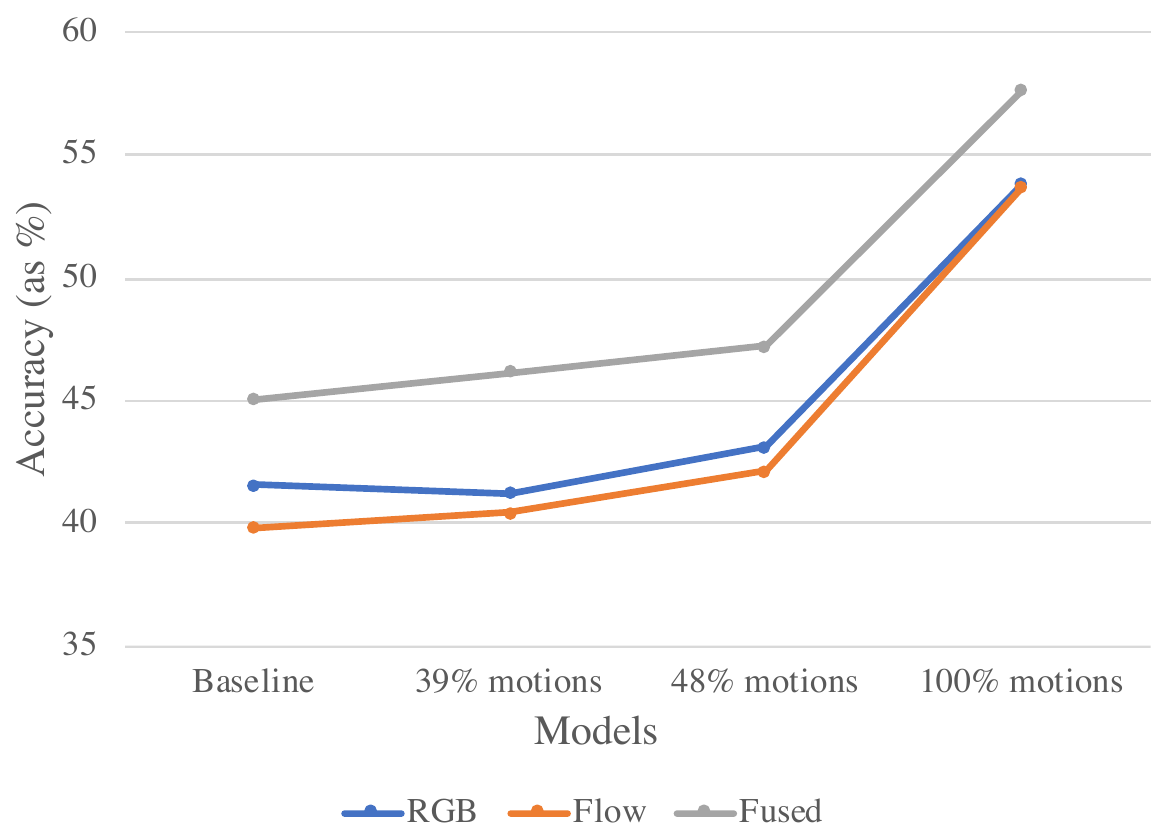}
  \caption{
    Top-1 verb classification accuracy on validation set based on accuracy of
    motion embedding features.
  }
  \label{fig:motions_verb_classsif}
\end{figure}

The results from Table \ref{table:verb_clasification_small} demonstrate that motion embedding features benefit the verb classifier, but the improvements had insignificant margins.
We believe that this is due to the relatively low motion embedding accuracy, which was caused by a small training set.
Therefore, we also evaluate our model on the validation set, which has ground truth motion codes, allowing us to use them as the motion features directly.
Table \ref{table:verb_clasification_validation} shows the top-1 verb classification accuracy results on the validation set.
As previously, we can see that model with motion features from vanilla motion embedding model $\hat{V}(V_{x}, M_{x})$ is slightly better than the baseline,
while $\hat{V}(V_{x}, M_{x,z})$ makes improvements more significant.
However, if we have ground-truth motion codes (i.e., 100\% accurate), its accuracy
becomes even higher than $V_{x,z}$, which uses ground truth nouns.
In other words, the features from the motion codes are more visually valuable
to the verb classifier than the knowledge about the object-in-action.
Figure \ref{fig:motions_verb_classsif} illustrates the trend of the verb
classification accuracy based on the accuracy of motion features.

\begin{figure*}
  \centering
  \includegraphics[width=0.9\textwidth]{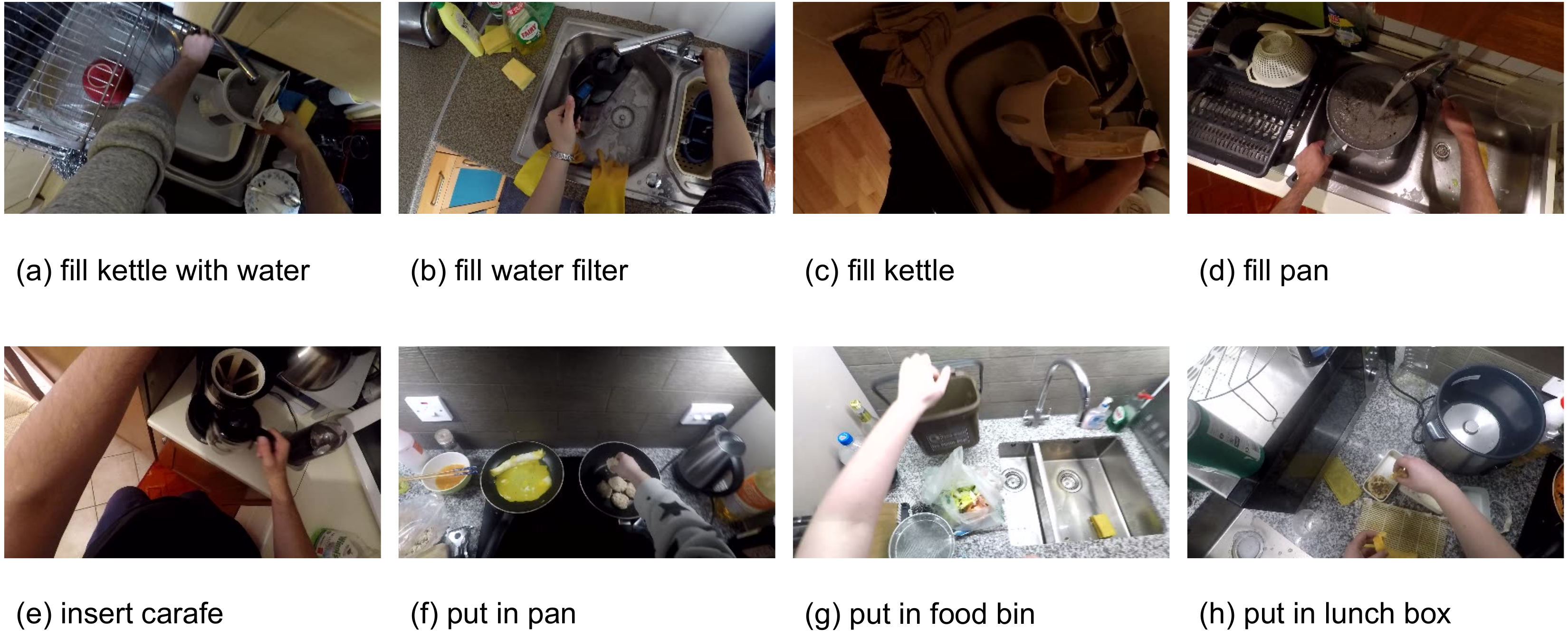}
  \caption{
    Videos from ``fill" (top row) and ``insert" (bottom row) classes that were 
    incorrectly classified by our model.
  }
  \label{fig:examples}
\end{figure*}

\begin{figure}
  \centering
  \includegraphics[trim={2.6cm 3cm 4cm 4cm},clip,width=\columnwidth]{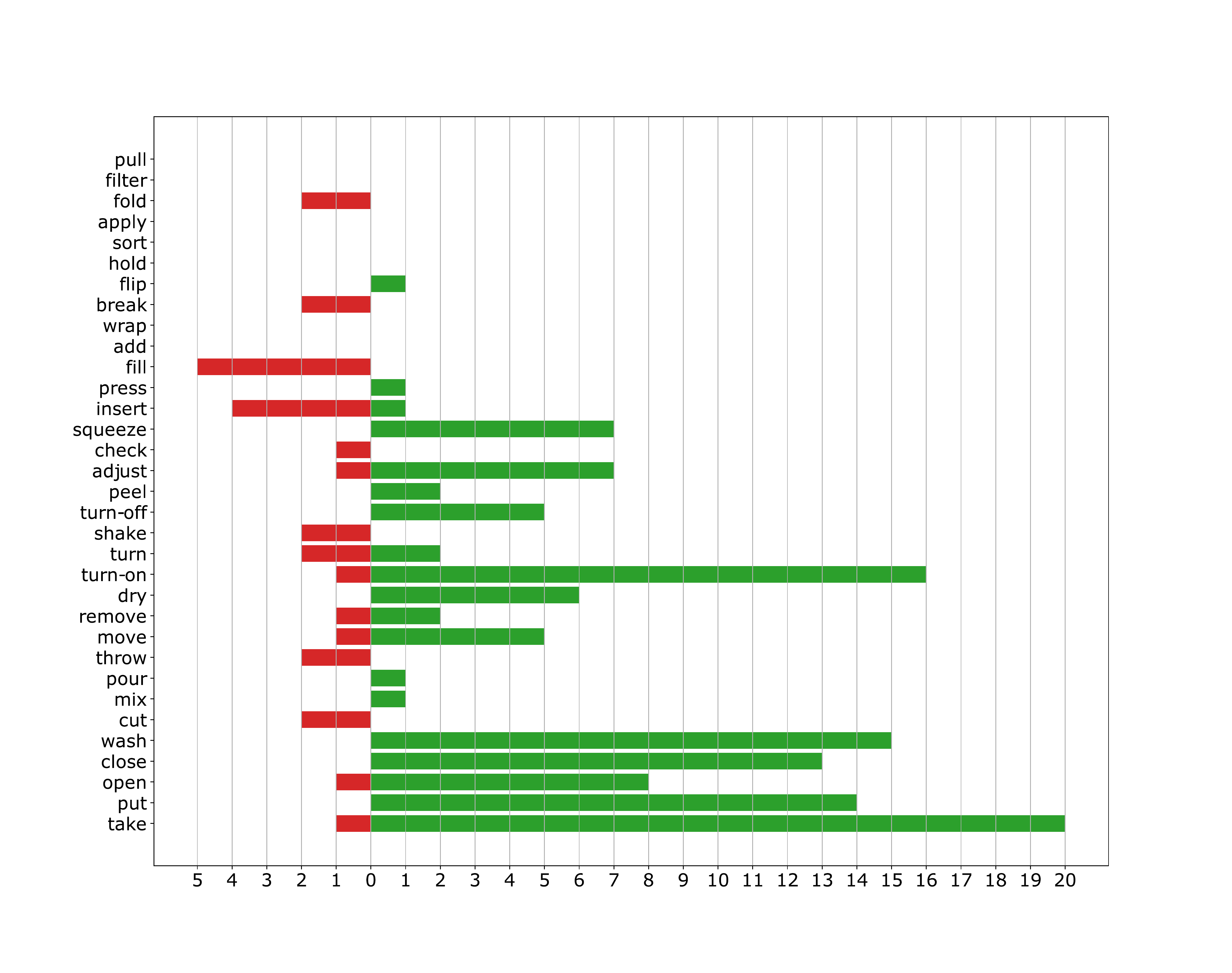}
  \caption{
    Number of videos that were correctly classified by 
    $\hat{V}(V_{x},\Bar{M}_{x})$ but incorrectly classified by $V_{x}$ (green) 
    and vice versa (red).
  }
  \label{fig:true_motion_vs_baseline}
\end{figure}

From Tables \ref{table:verb_clasification_small} and \ref{table:verb_clasification_validation}, 
we can observe that, in most cases, models that use optical flow frames benefit more from using motion codes than their RGB counterparts. 
In Table \ref{table:verb_clasification_small}, the first baseline $V_{x}$ classifies verbs more accurately when it uses RGB frames as opposed to using optical flow frames. 
After applying motions predicted by $M_{x}$, we can see that the optical flow model improved more than the RGB model, although still giving lower overall accuracy. 
However, after applying an improved motion code embedding $M_{x,z}$, the performance of the flow model was better than the RGB model. 
In the baseline model with nouns $V_{x,z}$, RGB and flow models have very close performances, but model $\hat{V}(V_{x,z},M_{x,z})$ performs noticeably better with flow frames.
Table \ref{tab:motion_prediction} also supports that motions predicted from flow frames are consistently more accurate than from RGB frames. 
On validation set, both baseline verb classifiers, $V_{x}$ and $V_{x,z}$, perform significantly better on RGB modality, but that distance gets reduced with motion embedding features and approaches its minimum when using ground truth motion codes as shown in Table \ref{table:verb_clasification_validation}. 

To summarize, as we elaborated in Section \ref{sec:related-work}, in 
\cite{varol2017long}, the authors emphasized the importance of motion feature estimation for action recognition, such as more accurate optical flow computation. 
Our observations corroborate this viewpoint as we demonstrated a high correlation between our motion codes and optical flow vectors. 
In addition, the motion codes demonstrate that they provide more visually informative data in comparison with nouns.
We acknowledge that in real scenarios, the model will have access to neither ground truth nouns nor motion codes.
However, these observations show that if we improve motion embedding accuracy (e.g. by increasing the training dataset size), we will likely achieve more significant improvements in verb classification.
Therefore, we believe that our motion code embedding model has the potential to be a significant add-on feature extractor for action recognition, especially in
providing explainable motion features.

\subsection{Qualitative Results}

Figure \ref{fig:true_motion_vs_baseline} shows the number of videos that were 
incorrectly classified by the baseline model and were corrected by our model 
that uses ground-truth motion codes. 
It also shows videos that were misclassified by our model but were correct on the baseline. 
We can see that most of the classes benefit from using ground truth motion codes, although with some exceptions. 
Five videos of class ``fill" were predicted correctly with the baseline model 
and were misclassified by our model. 
Instead, they were classified as ``put" (2 instances) or ``take" (3 instances). 
By watching those videos, we noticed that motions executed on them were indeed closer to put or take. 
All of them illustrate a person holding a kettle, water filter, or pan under the tap while the tap is filling those containers with water.
Meanwhile, the most salient motion in all videos was the actor bringing and placing the container under the tap.
Figure \ref{fig:examples} demonstrates sample frames of those videos with their annotated narrations. 

Another class that suffered from our model is ``insert" with four misclassified videos. 
What is interesting about those videos is that three out of four of them were narrated as ``put-in" action, while the class key is called ``insert". 
Our model classified these videos as ``put" class, and the videos themselves demonstrate how the actor puts the passive object on something, as shown on the bottom row of Figure \ref{fig:examples}.
Therefore, we consider this misclassification to be caused by grouping verbs into coarse-grained classes, since in 3 out of 4 misclassified videos, the actual verbs 
that were narrated by the annotators match the class that our model predicted.

In short, the results shown in Figure \ref{fig:true_motion_vs_baseline} and the 
analysis of videos that were misclassified by our model confirm that motion 
codes provide additional robustness to action recognition from the motion 
mechanics perspective.
Had more fine-grained labels been used, the predicted labels would more 
adequately describe the action taking place. 

\section{Conclusion and Future Work}

We have defined motion code embedding using motion taxonomy and trained a model that can project videos to the motion embedding space. 
We also showed how such a kind of model could be used along with a state-of-the-art verb classification model and improve its performance. 
We acknowledge that the presented method of augmenting verb classifiers with motion code embedding is simple and more sophisticated methods could be adopted in the future. 
For instance, in determining contact type and duration, we can use visual features such as bounding boxes drawn over objects-in-action derived from object detection models to calculate the duration of intersection between active and passive object boundaries.
The next step of this work could be unfolding the verb labels of the EPIC-KITCHENS dataset and make them more fine-grained, as the current verb classes are arguably quite ambiguous. 
One method could embed the actual verbs that were narrated by the annotator (i.e. not labels) with our word embedding model and train the model to find the nearest neighbor vector. 
We would also want to annotate more training videos to improve the accuracy of the motion code embedding model, as our current training set is 10 times smaller than the entire EPIC-KITCHENS dataset. 
We will also attempt to annotate videos from other datasets and use them to test our motion code embedding that was trained on EPIC-KITCHENS to see how our model can be generalized to other datasets.

\section*{Acknowledgements}
This material is based upon work supported by the National Science Foundation under Grant Nos. 1812933 and 1910040.

\bibliographystyle{IEEEtran}
\bibliography{bibliography}

\end{document}